\documentclass{article}

\usepackage[dblblindworkshop, final]{neurips_2025}

\usepackage[utf8]{inputenc} 
\usepackage[T1]{fontenc}    
\usepackage{natbib} 
\bibpunct{[}{]}{,}{n}{}{,}
\usepackage{hyperref}       
\usepackage{url}            
\usepackage{booktabs}       
\usepackage{amsfonts}       
\usepackage{nicefrac}       
\usepackage{microtype}      
\usepackage[table,xcdraw]{xcolor}
\usepackage{amsmath}
\usepackage{graphicx}
\usepackage{subcaption}

\title{Self-Evaluating LLMs for Multi-Step Tasks: Stepwise Confidence Estimation for Failure Detection}

\author{%
  Vaibhav Mavi \\
  Dyania Health\\
  \texttt{vaibhav@dyaniahealth.com} \\
  \And
  Shubh Jaroria \\
  Dyania Health \\
  \texttt{shubh@dyaniahealth.com} \\  \And
  Weiqi Sun \\
  Dyania Health \\
  \texttt{weiqi@dyaniahealth.com} \\
}

\begin{document}

\maketitle

\begin{abstract}
Reliability and failure detection of large language models (LLMs) is critical for their deployment in high-stakes, multi-step reasoning tasks. Prior work explores confidence estimation for self-evaluating \textit{LLM-scorer systems}, with confidence scorers estimating the likelihood of errors in LLM responses. However, most methods focus on single-step outputs and overlook the challenges of multi-step reasoning. In this work, we extend self-evaluation techniques to multi-step tasks, testing two intuitive approaches: holistic scoring and step-by-step scoring. Using two multi-step benchmark datasets, we show that stepwise evaluation generally outperforms holistic scoring in detecting potential errors, with up to 15\% relative increase in AUC-ROC. Our findings demonstrate that self-evaluating LLM systems provide meaningful confidence estimates in complex reasoning, improving their trustworthiness and providing a practical framework for failure detection.
\end{abstract}

\section{Introduction}\label{sec:intro}
Large language model (LLM) agents are increasingly deployed in complex applications such as task-planning \cite{survey-planning}, dialog systems \cite{survey-dialog}, collaborative problem-solving \cite{survey-collab} and multi-hop question answering \cite{survey-mhqa} where detecting errors and failures is a critical challenge. A common strategy for detecting failures is to extend the system with a self-evaluation component, where either the agent itself or an auxiliary evaluator assigns a confidence score to the response \cite{survey}.
\par
Failure detection through confidence estimation has been extensively studied in single-step prediction tasks \cite{survey, elicit, blackbox, activations1}, but its role in multi-step reasoning remains largely underexplored. Multi-step interactions pose unique challenges: reasoning chains can be arbitrarily long, errors may occur at any step, and later steps often depend on earlier ones. Consequently, direct application of existing methods often fails to identify errors in a multi-step task reliably. For example, self-certainty \cite{self-certainty} directly applied to CoQA (Conversational Question Answering) \cite{coqa} yields poor performance (AUC-ROC 0.523, FPR@0.9 recall 0.95).
\par
However, a trivial extension of detecting errors after each step improves the performance substantially (AUC-ROC \textbf{0.849}, FPR@0.9 recall \textbf{0.374}). This observation raises a key question: Should confidence estimation methods in multi-step tasks evaluate responses: i) after each reasoning step, enabling fine-grained error-detection, or ii) holistically, considering the final answer in full context?

We systematically investigate this question across two representative settings: tool-enhanced reasoning and LLM–user dialog. Our experiments reveal that step-level evaluation often provides superior error detection, though holistic evaluation can still be advantageous in certain contexts.


\begin{figure}
  \centering
  \begin{subfigure}{0.53\linewidth}
      \centering
      \includegraphics[width=\linewidth]{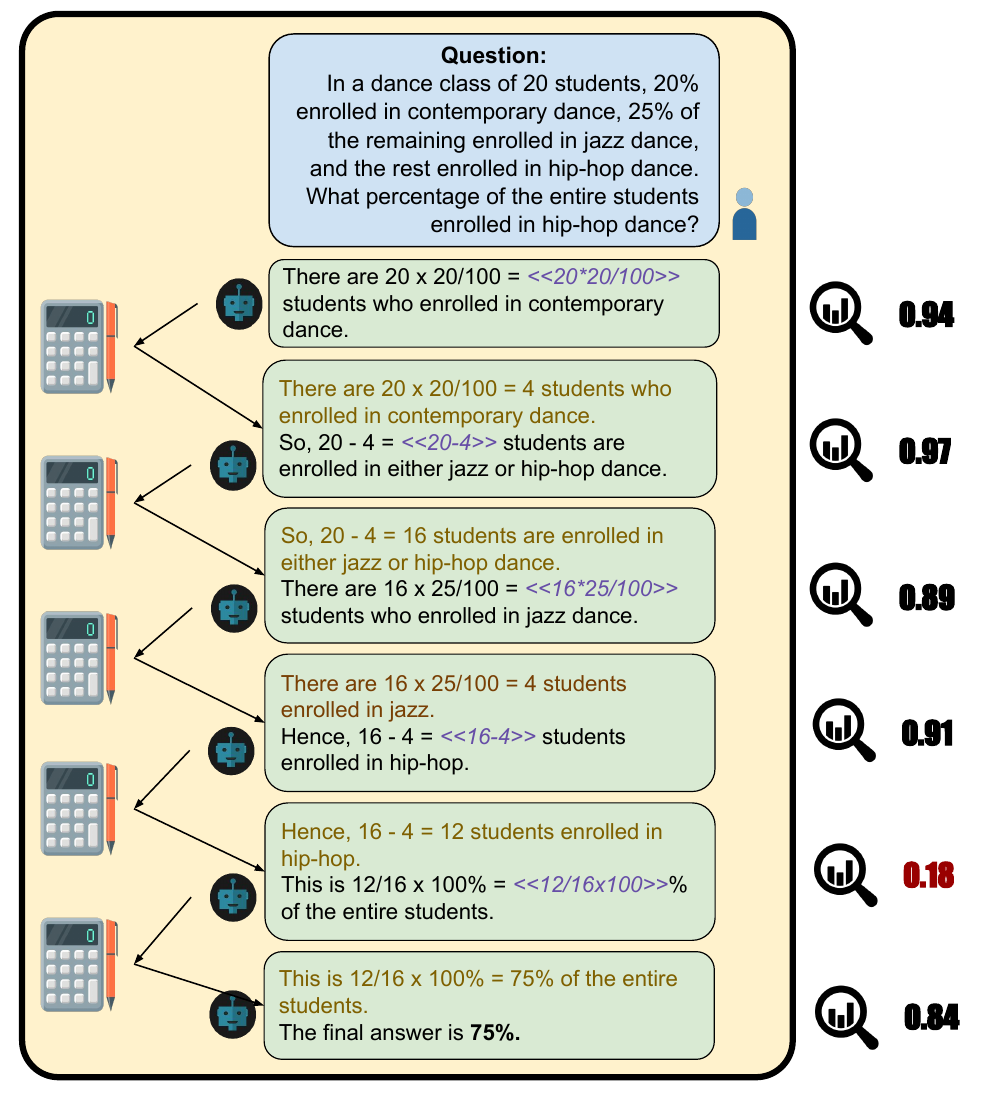}
      \caption{step-level confidence scoring setup with GSM8K}
      \label{fig:gsm}
    \end{subfigure}
    \hfill
    \begin{subfigure}{0.45\linewidth}
        \centering
        \includegraphics[width=\linewidth]{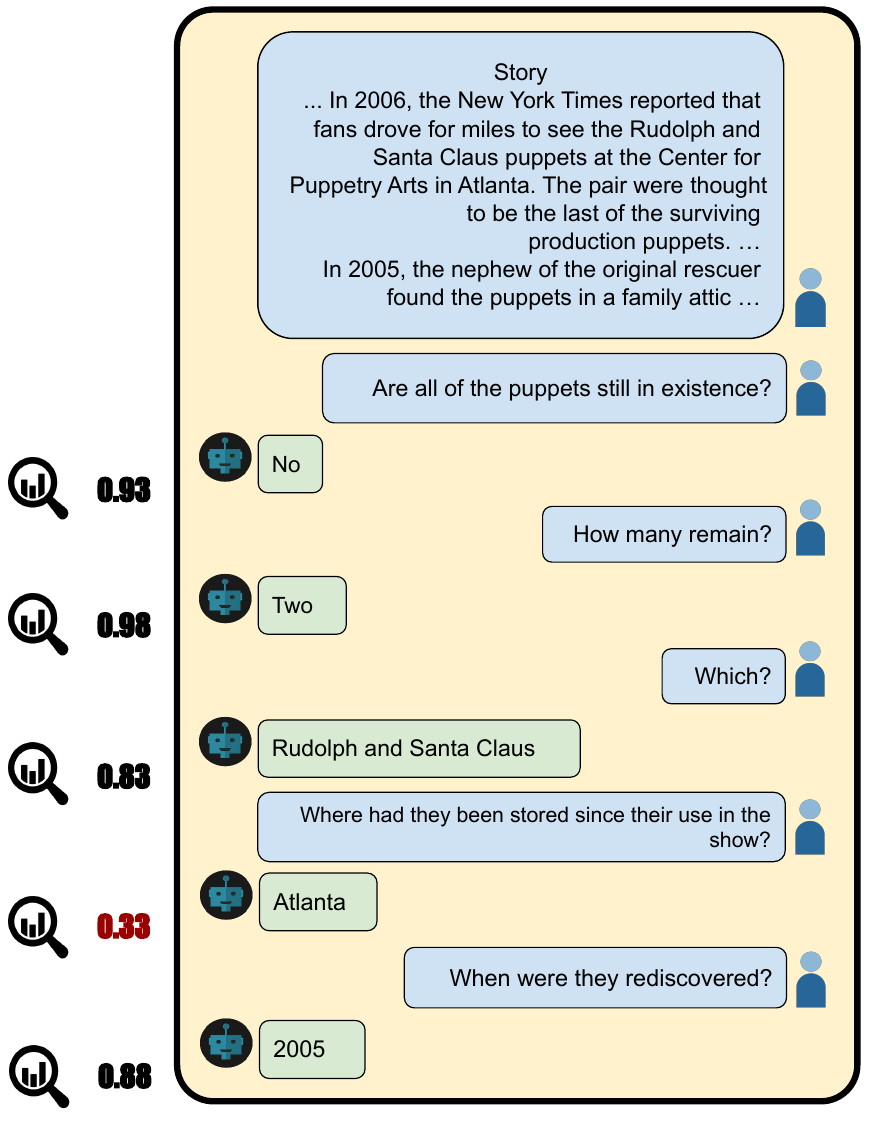}
        \caption{step-level confidence scoring setup with CoQA}
        \label{fig:coqa}
    \end{subfigure}
\end{figure}

\section{Related Work}
Prior work on confidence estimation for LLMs can be broadly divided into black-box and white-box approaches.
Black-box methods assume no access to the underlying model and often rely on prompting-based strategies such as self-reflection, self-consistency, or generating multiple candidate answers \cite{self-reflection, self-consistency, mutliple-answers, preferknow, mostly-know, steerconf-verbal}. Another line of work employs external evaluators to assess responses post hoc, using features such as similarity measures or structured scoring models \cite{graph, fine-grained, blackbox, similarity-features}.

White-box methods, in contrast, target open-source models where full access to parameters and activations is possible. These techniques include fine-tuning models to improve self-evaluation abilities \cite{past-experience, doubt-rl}, leveraging log-probabilities for calibrated confidence estimation \cite{dst, multicalibration, self-certainty}, and training regression models over hidden states to predict correctness \cite{activations1, activations2, activations3}.

\citeauthor{survey} \cite{survey} provide a comprehensive survey of these approaches, organizing calibration techniques across settings and highlighting key limitations. However, the vast majority of existing work studies single-step tasks, where the model outputs a single response to a single query. Extending these methods to multi-step reasoning remains largely unaddressed.

\section{Problem Definition}
We define failure detection as the task of estimating the probability $p \in [0,1]$ that an agent’s ($A$) response $R$ to a given input $I$ is incorrect: $p = \mathcal{F}(R \mid$ I).
\par
Multi-step interactions require a more general formulation: the input is defined as $I = (C, Q)$ where $C$ denotes the initial context and $Q$ is the sequence of queries. The LLM agent produces a sequence of responses $R$ with interaction length $n = \vert Q \vert = \vert R \vert$. We consider the following two adaptations:
\par
\textbf{Response-level scoring}: Treat the queries and responses in all the steps as a single sequence and assign one confidence score to the whole solution. This holistic approach captures global coherence.
\begin{equation}\label{eq:whole}
    p = \mathcal{S}_{whole}(R_{[1:n]} \mid C, Q_{[1:n]})
\end{equation}
\par
\textbf{Step-level scoring}: The response $R_i$ at a given step $i$ is dependent on the prior queries and responses and scoring it requires all of the previous context ($C, Q_{[1:i]}, R_{[1:i-1]}$). Accordingly, we assign a separate score to each response $R_i$, conditioned on the previous queries and responses. 

\begin{equation}\label{eq:step}
    p_i = \mathcal{F}_{step}(R_i \mid C, Q_{[1:i]}, R_{[1:i-1]}) \\
\end{equation}
\par
If any individual score exceeds the threshold, the entire response can be flagged as potentially incorrect. This can be achieved by using $p=min(\{p_i\}_{i=1}^n)$.

\section{Data}
\subsection{Agent Inputs}
To test error detection, we focus on tasks where correctness can be objectively defined at each step. Accordingly, we select the following datasets.
\par
\textbf{GSM8K} (Grade School Math - 8K) \cite{gsm8k} is a collection of grade-school math word problems that require multi-step reasoning and computation. At each step, the agent generates an intermediate formula, queries an expression evaluator, and incorporates the tool's response into subsequent steps (Figure \ref{fig:gsm} ). Problems in the GSM8K test set require an average of 5.1 steps. 
\par
\textbf{CoQA} (Conversational Question Answering) \cite{coqa} contains over 127,000 question–answer pairs spanning 8,000 conversations that are context-grounded, with later questions often depending on previous queries and answers (Figure \ref{fig:coqa}). Conversations have an average of 13.5 steps.
\par
\textbf{Responses:}\\
For both tasks, we fine-tune Llama-3.2-11B-Instruct \cite{llama32} for two epochs. Because several confidence scoring methods require training, we hold out subsets of training and test splits for confidence estimation, while using the remainder to train the LLM agent. 
For GSM8K specifically, we use two sets of labels: Answer labels — assess whether the final answer matches the ground truth, and Reasoning labels — assess whether each intermediate reasoning step is correct. Further details on response labels and accuracy are included in Appendix \ref{app:train}.

\section{Experiments}
\subsection{Confidence Estimation Methods}
We evaluate several confidence scoring methods, under both formulations response-level (Equation \ref{eq:whole}) and step-level (Equation \ref{eq:step}). For methods requiring training, we use the instruction-tuned Llama-3.2-11B as the base model, replacing its generation head with a regression head for classification objectives. Details on each algorithm are mentioned in the Appendix. \ref{app:methods}

\subsection{Evaluation Metrics}
We frame error detection as a binary classification task and report the following metrics: \par \textbf{AUC-ROC} (Area Under the Receiver Operating Characteristic Curve): Measures how well the model separates correct from incorrect responses across thresholds. Higher is better.
\par \textbf{FPR@0.9 Recall}: Since the goal is to reliably flag potentially incorrect responses while minimizing false alarms, we measure the false positive rate (FPR) of the model at a threshold where it identifies the incorrect responses with at least $0.9$ recall. Some approaches fail to reach the target recall without trivially classifying all responses as incorrect. In these cases, we report FPR@0.9 recall as $1$ and additionally report the maximum achievable recall.

\subsection{Results}

\textbf{Failure detection in multi-step interactions:}
For both tasks, the best performing methods achieve an AUC-ROC of $0.9$ and a recall of $0.9$ with FPR below one-third. Across techniques, regression model performs the best for both tasks. Interestingly, preference-based reward models perform poorly, suggesting that PRMs are better suited for ranking responses by quality, rather than tasks that have objective correctness labels \cite{reward2}.

\textbf{Performance across granularity and task:}
For CoQA, step-level scoring significantly outperforms response-level scoring across all methods. For GSM8K, the difference is smaller and trends are less consistent. 
Notably, the step-level performance of self-certainty is significantly worse, likely due to tool interactions that alter the agent’s responses at each step, thereby distorting the logits. This degradation is not observed for the activations-based regressor, since it only relies on hidden states from the final token.

Most techniques perform better on CoQA than GSM8K, suggesting that reasoning-intensive math problems might be more challenging for evaluators than context-grounded QA. Interestingly, GPT-4.1-mini shows significantly improved performance on GSM8K, reflecting its superior reasoning ability.

\begin{table}
    \centering
    \small
    \caption{Evaluation results of different techniques on GSM8K and CoQA. An FPR@0.9 Recall of 1.0 (mr: \textit{x}) means that the recall does not exceed \textit{x} without flagging everything as low confidence.}
    \begin{tabular}{ll|cc|cc}
        \toprule
        \multicolumn{2}{c}{} & \multicolumn{2}{c}{\textbf{GSM8K}} & \multicolumn{2}{c}{\textbf{CoQA}} \\
        \toprule
        \textbf{Technique} & \textbf{granularity} & {AUC} ($\uparrow$) & {FPR@0.9 rec} ($\downarrow$) & {AUC} ($\uparrow$) & {FPR@0.9 rec} ($\downarrow$) \\
        \midrule

         & response & 0.556 & 1.0 (mr: 0.13) & 0.502 & 1.0 (mr: 0.77) \\
        Self-verbalized & step & 0.546 (-0.2\%) & 1.0 (mr: 0.10) & 0.624 (+24\%) & 0.587 \\
        \midrule

         & response & 0.586 & 1.0 (mr: 0.52) & 0.522 & 1.0 (mr: 0.73) \\
        Llama-3.2-11B & step & 0.676 (+15\%) & 1.0 (mr: 0.75) & 0.613 (+12\%) & 0.81 \\
        \midrule
        
         & response & 0.880  & 1.0 (mr: 0.81) & 0.548 & 0.88 \\
        GPT-4.1-mini & step & 0.670 (-24\%) & 1.0 (mr: 0.48) & 0.665 (+21\%) & 0.476 \\
        \midrule

         & response & 0.843 & 0.441 & 0.689 & 0.732 \\
        Regression & step & \textbf{0.907} (+7\%) & \textbf{0.314} & \textbf{0.952} (+38\%) & \textbf{0.169} \\
        \midrule
        
         & response & 0.450  & 0.928 & 0.381 & 1.0 (mr: 0.57) \\
        PRM & step & - & - & 0.493 (+30\%) & 0.887 \\
        \midrule
        
         & response & 0.649 & 0.812 & 0.523 & 0.95 \\
        Self-certainty & step & 0.395 (-40\%) & 0.945 & 0.849 (+62\%) & 0.374 \\
        \midrule
        
         & response & 0.608 & 1 (mr: 0.77) & 0.792 & 0.643 \\
        Activations & query & 0.750 (+23\%) & 0.647 & 0.919 (+16\%) & \textbf{0.169} \\
        \bottomrule

    \end{tabular}
    \label{tab:gsm}
\end{table}

\textbf{Relation to final answer accuracy:}
For GSM8K, the agent reached the correct answer despite flawed intermediate reasoning in \textbf{60/879} test cases (Figure \ref{fig:step}).
Table \ref{tab:gsm-ans} shows that step-level performance of all methods against final answers is slightly lower, while the response-level performance improves. This is expected since step-level scoring penalizes intermediate mistakes more strongly, while response-level scoring focuses on the overall outcome.

Identifying cases where the agent reaches the correct answer through flawed reasoning is crucial for trustworthy deployment. Table \ref{tab:step} shows that for most methods, step-level scoring is more effective at detecting such cases.

\begin{table}
    \centering
    \small
    \caption{Recall for cases with incorrect reasoning steps but correct answer. Higher the better}
    \begin{tabular}{l c c c c c c}
         \toprule
          & Self-eval & Llama-3.2-11B & GPT-4.1-mini & Regression & Self-certainty & Activation \\
         \midrule
         response & 0.05 & 0.133 & 0.50 & 0.367 & 0.133 & 0.167 \\
         step & 0 & 0.40 & 0.30 & \textbf{0.60} & 0.217 & 0.267 \\
         \bottomrule
    \end{tabular}
    \label{tab:step}
\end{table}

\textbf{Case study on real world data}
We also test the effectiveness of this approach on a private dataset with real clinical notes and questions.
Consistent with the analysis on public datasets, a regression model generating step-level scores achieves the best performance with \textbf{AUC-ROC} of $\mathbf{=0.940}$ and \textbf{FPR@0.9 rec} $\mathbf{=0.152}$. We include further details in Appendix \ref{app:private}.

\section{Conclusion}
We extended confidence estimation to multi-step tasks in dialogue and tool-assisted reasoning, where maintaining consistency across steps is especially challenging. Through experiments on two multi-step tasks, we find that step-level scoring, though harder to implement in some cases, generally improves error detection and reveals when correct answers emerge from faulty reasoning. Our study highlights the limits of current methods and provides a basis for developing confidence estimators better suited to multi-step reasoning.

\bibliographystyle{plainnat}  
\bibliography{references}

@misc{teacher-forcing,
      title={Professor Forcing: A New Algorithm for Training Recurrent Networks}, 
      author={Alex Lamb and Anirudh Goyal and Ying Zhang and Saizheng Zhang and Aaron Courville and Yoshua Bengio},
      year={2016},
      eprint={1610.09038},
      archivePrefix={arXiv},
      primaryClass={stat.ML},
      url={https://arxiv.org/abs/1610.09038}, 
}

@misc{coqa,
      title={CoQA: A Conversational Question Answering Challenge}, 
      author={Siva Reddy and Danqi Chen and Christopher D. Manning},
      year={2019},
      eprint={1808.07042},
      archivePrefix={arXiv},
      primaryClass={cs.CL},
      url={https://arxiv.org/abs/1808.07042}, 
}

@misc{graph,
      title={Graph-based Confidence Calibration for Large Language Models}, 
      author={Yukun Li and Sijia Wang and Lifu Huang and Li-Ping Liu},
      year={2025},
      eprint={2411.02454},
      archivePrefix={arXiv},
      primaryClass={cs.CL},
      url={https://arxiv.org/abs/2411.02454}, 
}

@misc{elicit,
      title={Can LLMs Express Their Uncertainty? An Empirical Evaluation of Confidence Elicitation in LLMs}, 
      author={Miao Xiong and Zhiyuan Hu and Xinyang Lu and Yifei Li and Jie Fu and Junxian He and Bryan Hooi},
      year={2024},
      eprint={2306.13063},
      archivePrefix={arXiv},
      primaryClass={cs.CL},
      url={https://arxiv.org/abs/2306.13063}, 
}

@inproceedings{self-reflection,
  title        = {Reflexion: Language Agents with Verbal Reinforcement Learning},
  author       = {Noah Shinn and Federico Cassano and Ashwin Gopinath and Karthik Narasimhan and Shunyu Yao},
  booktitle    = {Advances in Neural Information Processing Systems 36 – 37th Conference on Neural Information Processing Systems, NeurIPS 2023},
  editor       = {A. Oh and T. Neumann and A. Globerson and K. Saenko and M. Hardt and S. Levine},
  publisher    = {Neural Information Processing Systems Foundation},
  year         = {2023},
  series       = {Advances in Neural Information Processing Systems},
  eprint       = {2303.11366},
  archivePrefix = {arXiv},
  primaryClass = {cs.AI}
}

@misc{preferknow,
      title={Language Models Prefer What They Know: Relative Confidence Estimation via Confidence Preferences}, 
      author={Vaishnavi Shrivastava and Ananya Kumar and Percy Liang},
      year={2025},
      eprint={2502.01126},
      archivePrefix={arXiv},
      primaryClass={cs.CL},
      url={https://arxiv.org/abs/2502.01126}, 
}

@misc{fine-grained,
      title={Mind the Generation Process: Fine-Grained Confidence Estimation During LLM Generation}, 
      author={Jinyi Han and Tingyun Li and Shisong Chen and Jie Shi and Xinyi Wang and Guanglei Yue and Jiaqing Liang and Xin Lin and Liqian Wen and Zulong Chen and Yanghua Xiao},
      year={2025},
      eprint={2508.12040},
      archivePrefix={arXiv},
      primaryClass={cs.CL},
      url={https://arxiv.org/abs/2508.12040}, 
}

@misc{activations3,
      title={The Internal State of an LLM Knows When It's Lying}, 
      author={Amos Azaria and Tom Mitchell},
      year={2023},
      eprint={2304.13734},
      archivePrefix={arXiv},
      primaryClass={cs.CL},
      url={https://arxiv.org/abs/2304.13734}, 
}

@misc{activations2,
      title={Discovering Latent Knowledge in Language Models Without Supervision}, 
      author={Collin Burns and Haotian Ye and Dan Klein and Jacob Steinhardt},
      year={2024},
      eprint={2212.03827},
      archivePrefix={arXiv},
      primaryClass={cs.CL},
      url={https://arxiv.org/abs/2212.03827}, 
}

@misc{mostly-know,
      title={Language Models (Mostly) Know What They Know}, 
      author={Saurav Kadavath and Tom Conerly and Amanda Askell and Tom Henighan and Dawn Drain and Ethan Perez and Nicholas Schiefer and Zac Hatfield-Dodds and Nova DasSarma and Eli Tran-Johnson and Scott Johnston and Sheer El-Showk and Andy Jones and Nelson Elhage and Tristan Hume and Anna Chen and Yuntao Bai and Sam Bowman and Stanislav Fort and Deep Ganguli and Danny Hernandez and Josh Jacobson and Jackson Kernion and Shauna Kravec and Liane Lovitt and Kamal Ndousse and Catherine Olsson and Sam Ringer and Dario Amodei and Tom Brown and Jack Clark and Nicholas Joseph and Ben Mann and Sam McCandlish and Chris Olah and Jared Kaplan},
      year={2022},
      eprint={2207.05221},
      archivePrefix={arXiv},
      primaryClass={cs.CL},
      url={https://arxiv.org/abs/2207.05221}, 
}

@misc{dst,
      title={Confidence Estimation for LLM-Based Dialogue State Tracking}, 
      author={Yi-Jyun Sun and Suvodip Dey and Dilek Hakkani-Tur and Gokhan Tur},
      year={2024},
      eprint={2409.09629},
      archivePrefix={arXiv},
      primaryClass={cs.CL},
      url={https://arxiv.org/abs/2409.09629}, 
}

@inproceedings{activations1,
    title = "{I}nternal{I}nspector $I^2$: Robust Confidence Estimation in {LLM}s through Internal States",
    author = "Beigi, Mohammad  and
      Shen, Ying  and
      Yang, Runing  and
      Lin, Zihao  and
      Wang, Qifan  and
      Mohan, Ankith  and
      He, Jianfeng  and
      Jin, Ming  and
      Lu, Chang-Tien  and
      Huang, Lifu",
    editor = "Al-Onaizan, Yaser  and
      Bansal, Mohit  and
      Chen, Yun-Nung",
    booktitle = "Findings of the Association for Computational Linguistics: EMNLP 2024",
    month = nov,
    year = "2024",
    address = "Miami, Florida, USA",
    publisher = "Association for Computational Linguistics",
    url = "https://aclanthology.org/2024.findings-emnlp.751/",
    doi = "10.18653/v1/2024.findings-emnlp.751",
    pages = "12847--12865",
}

@misc{self-consistency,
      title={Self-Consistency Improves Chain of Thought Reasoning in Language Models}, 
      author={Xuezhi Wang and Jason Wei and Dale Schuurmans and Quoc Le and Ed Chi and Sharan Narang and Aakanksha Chowdhery and Denny Zhou},
      year={2023},
      eprint={2203.11171},
      archivePrefix={arXiv},
      primaryClass={cs.CL},
      url={https://arxiv.org/abs/2203.11171}, 
}

@inproceedings{survey,
    title ={"A Survey of Confidence Estimation and Calibration in Large Language Models"},
    author = {Geng, Jiahui  and
      Cai, Fengyu  and
      Wang, Yuxia  and
      Koeppl, Heinz  and
      Nakov, Preslav  and
      Gurevych, Iryna},
    editor = {"Duh, Kevin  and
      Gomez, Helena  and
      Bethard, Steven"},
    booktitle = {"Proceedings of the 2024 Conference of the North American Chapter of the Association for Computational Linguistics: Human Language Technologies (Volume 1: Long Papers)"},
    month = {jun},
    year = {2024},
    address = {"Mexico City, Mexico"},
    publisher = {"Association for Computational Linguistics"},
    url = {"https://aclanthology.org/2024.naacl-long.366/"},
    doi = {"10.18653/v1/2024.naacl-long.366"},
    pages = {"6577--6595"},
}

@misc{self-certainty,
      title={Scalable Best-of-N Selection for Large Language Models via Self-Certainty}, 
      author={Zhewei Kang and Xuandong Zhao and Dawn Song},
      year={2025},
      eprint={2502.18581},
      archivePrefix={arXiv},
      primaryClass={cs.CL},
      url={https://arxiv.org/abs/2502.18581}, 
}

@inproceedings{similarity-features,
    title={Assessing Confidence in Large Language Models by Classifying Task Correctness using Similarity Features},
    author={Debarun Bhattacharjya and Balaji Ganesan and Junkyu Lee and Radu Marinescu},
    booktitle={ICLR Workshop: Quantify Uncertainty and Hallucination in Foundation Models: The Next Frontier in Reliable AI},
    year={2025},
    url={https://openreview.net/forum?id=DirbdPbGhv}
    }

@misc{mutliple-answers,
      title={Think Twice Before Trusting: Self-Detection for Large Language Models through Comprehensive Answer Reflection}, 
      author={Moxin Li and Wenjie Wang and Fuli Feng and Fengbin Zhu and Qifan Wang and Tat-Seng Chua},
      year={2024},
      eprint={2403.09972},
      archivePrefix={arXiv},
      primaryClass={cs.CL},
      url={https://arxiv.org/abs/2403.09972}, 
}

@misc{blackbox,
      title={Large Language Model Confidence Estimation via Black-Box Access}, 
      author={Tejaswini Pedapati and Amit Dhurandhar and Soumya Ghosh and Soham Dan and Prasanna Sattigeri},
      year={2025},
      eprint={2406.04370},
      archivePrefix={arXiv},
      primaryClass={cs.CL},
      url={https://arxiv.org/abs/2406.04370}, 
}

@misc{multicalibration,
      title={Multicalibration for Confidence Scoring in LLMs}, 
      author={Gianluca Detommaso and Martin Bertran and Riccardo Fogliato and Aaron Roth},
      year={2024},
      eprint={2404.04689},
      archivePrefix={arXiv},
      primaryClass={stat.ML},
      url={https://arxiv.org/abs/2404.04689}, 
}

@misc{past-experience,
      title={Enhancing Confidence Expression in Large Language Models Through Learning from Past Experience}, 
      author={Haixia Han and Tingyun Li and Shisong Chen and Jie Shi and Chengyu Du and Yanghua Xiao and Jiaqing Liang and Xin Lin},
      year={2024},
      eprint={2404.10315},
      archivePrefix={arXiv},
      primaryClass={cs.CL},
      url={https://arxiv.org/abs/2404.10315}, 
}

@misc{doubt-rl,
      title={Rewarding Doubt: A Reinforcement Learning Approach to Calibrated Confidence Expression of Large Language Models}, 
      author={Paul Stangel and David Bani-Harouni and Chantal Pellegrini and Ege Özsoy and Kamilia Zaripova and Matthias Keicher and Nassir Navab},
      year={2025},
      eprint={2503.02623},
      archivePrefix={arXiv},
      primaryClass={cs.CL},
      url={https://arxiv.org/abs/2503.02623}, 
}

@misc{steerconf-verbal,
      title={SteerConf: Steering LLMs for Confidence Elicitation}, 
      author={Ziang Zhou and Tianyuan Jin and Jieming Shi and Qing Li},
      year={2025},
      eprint={2503.02863},
      archivePrefix={arXiv},
      primaryClass={cs.CL},
      url={https://arxiv.org/abs/2503.02863}, 
}

@misc{gpt-4.1,
      title={GPT-4 Technical Report}, 
      author={OpenAI and Josh Achiam and ...},
      year={2024},
      eprint={2303.08774},
      archivePrefix={arXiv},
      primaryClass={cs.CL},
      url={https://arxiv.org/abs/2303.08774}, 
}

@misc{llama32,
      title={The Llama 3 Herd of Models}, 
      author={Aaron Grattafiori and Abhimanyu Dubey and ...},
      year={2024},
      eprint={2407.21783},
      archivePrefix={arXiv},
      primaryClass={cs.AI},
      url={https://arxiv.org/abs/2407.21783}, 
}

@misc{survey-dialog,
      title={A Survey on Recent Advances in LLM-Based Multi-turn Dialogue Systems}, 
      author={Zihao Yi and Jiarui Ouyang and Zhe Xu and Yuwen Liu and Tianhao Liao and Haohao Luo and Ying Shen},
      year={2025},
      eprint={2402.18013},
      archivePrefix={arXiv},
      primaryClass={cs.CL},
      url={https://arxiv.org/abs/2402.18013}, 
}

@article{survey-mhqa,
url = {http://dx.doi.org/10.1561/1500000102},
year = {2024},
volume = {17},
journal = {Foundations and Trends® in Information Retrieval},
title = {Multi-hop Question Answering},
doi = {10.1561/1500000102},
issn = {1554-0669},
number = {5},
pages = {457-586},
author = {Vaibhav Mavi and Anubhav Jangra and Adam Jatowt}
}

@article{
survey-planning,
author = {Wenshuo Zhai  and Jinzhi Liao  and Ziyang Chen  and Bolun Su  and Xiang Zhao },
title = {A Survey of Task Planning with Large Language Models},
journal = {Intelligent Computing},
volume = {4},
number = {},
pages = {0124},
year = {2025},
doi = {10.34133/icomputing.0124},
URL = {https://spj.science.org/doi/abs/10.34133/icomputing.0124},
eprint = {https://spj.science.org/doi/pdf/10.34133/icomputing.0124},
}

@misc{survey-collab,
      title={Multi-Agent Collaboration Mechanisms: A Survey of LLMs}, 
      author={Khanh-Tung Tran and Dung Dao and Minh-Duong Nguyen and Quoc-Viet Pham and Barry O'Sullivan and Hoang D. Nguyen},
      year={2025},
      eprint={2501.06322},
      archivePrefix={arXiv},
      primaryClass={cs.AI},
      url={https://arxiv.org/abs/2501.06322}, 
}

@misc{gsm8k,
      title={Training Verifiers to Solve Math Word Problems}, 
      author={Karl Cobbe and Vineet Kosaraju and Mohammad Bavarian and Mark Chen and Heewoo Jun and Lukasz Kaiser and Matthias Plappert and Jerry Tworek and Jacob Hilton and Reiichiro Nakano and Christopher Hesse and John Schulman},
      year={2021},
      eprint={2110.14168},
      archivePrefix={arXiv},
      primaryClass={cs.LG},
      url={https://arxiv.org/abs/2110.14168}, 
}

@misc{reward2,
      title={Reward Models Identify Consistency, Not Causality}, 
      author={Yuhui Xu and Hanze Dong and Lei Wang and Caiming Xiong and Junnan Li},
      year={2025},
      eprint={2502.14619},
      archivePrefix={arXiv},
      primaryClass={cs.LG},
      url={https://arxiv.org/abs/2502.14619}, 
}

\appendix

\section{Training details}\label{app:train}
\subsection{Confidence Scorers: Teacher Forcing}
For methods requiring supervised training, we adopt teacher forcing \cite{teacher-forcing}. During training, the model receives the gold history (i.e., corrected responses) when evaluating the next response. 
The learning objective is:
\begin{equation}\label{eq:train}
    p_i = \mathcal{F}(R_i, | C, Q_{[1:i]}, \hat{R}_{[1:i-1]}) = \mathbb{I}\{R_i \neq \hat{R}_i\}
\end{equation}
where $\hat{R}$ is the list of ground truth responses and $\mathbb{I}\{.\}$ is the indicator function. During inference, we do not assume access to the ground truth.
At inference time, however, no ground truth is available, and the evaluator must operate solely on the model’s predictions.

\section{Data preparation}
\subsection{Agent: Training and Inference}
For both tasks, we fine-tune Llama-3.2-11B-Instruct \cite{llama32} for two epochs. Because several confidence scoring methods require training, we hold out subsets of train and test splits for confidence estimation, while using the remainder to train the LLM agent. 
\par
\begin{figure}
    \centering
    \includegraphics[width=0.7\linewidth]{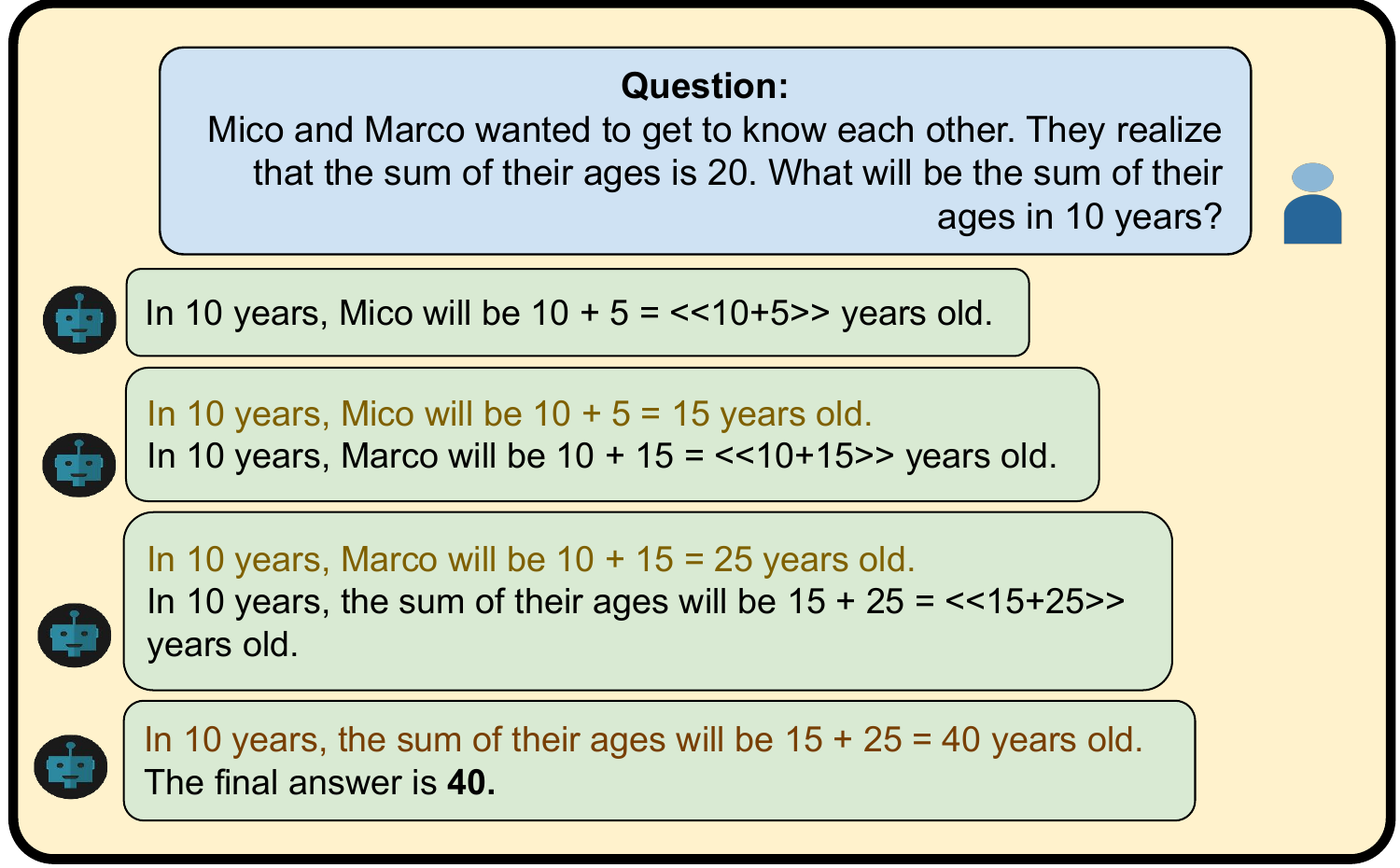}
    \caption{Case from GSM8K where the agent gets the answer correct through incorrect reasoning steps. The agent assumes the current ages of Mico and Marco to be $5$ and $15$ while the question does not mention it. The agent ends up getting to the correct answer nonetheless since it only concerns with the sum of their ages.}
    \label{fig:step}
\end{figure}

Performance varies across datasets and granularity:
- On CoQA, the agent achieves$\mathbf{81.2\%}$ step-level accuracy but only $\mathbf{16.1\%}$ response-level accuracy. The large gap is expected, since even a single incorrect step can propagate errors downstream.
- On GSM8K, the agent achieves $\mathbf{65.6\%}$ answer accuracy and $\mathbf{47.6\%}$ step-level accuracy. Here, answer accuracy is higher because the agent may arrive at correct final answers even if some intermediate steps are flawed (see Figure \ref{fig:step}).

\subsection{Labeling responses}\label{app:label}
We use GPT-5 to evaluate agent's responses against ground truth answers and intermediate steps, producing labels at both the step-level and response-level. To verify the label quality, we manually reviewed $100$ samples from each dataset. We found labeling accuracy above $96\%$ in both settings.

\section{Evaluated Confidence Estimation Methods}\label{app:methods}
\subsection{Black-box methods}
\subsubsection{Self-verbalized confidence}
The LLM agent is prompted to verbalize its confidence in its own response. For step-level scoring, the agent outputs the confidence score at the end of each step.

\subsubsection{Auxiliary evaluators}
External models assess the agent’s responses.
\par
\textbf{Pre-trained LLMs:}
Instruction-tuned LLMs are prompted to evaluate the agent's responses. We consider two evaluators: (a) Llama-3.2-11B (aligned with the agent’s base model), and (b) OpenAI’s GPT-4.1-mini \cite{gpt-4.1} (independent of the agent).
\par

\textbf{Regression model:}
We fine-tune the instruction tuned Llama-3.2-11B model with a sequence classification head to regress confidence scores in the range $[0,1]$. \par

\textbf{Preference-based reward model (PRM):}
We train a reward model on preference data, treating completions with correct answers as “chosen” and incorrect agent outputs as “rejected.” For GSM8K, multiple valid reasoning paths to solve the same problem make generating step-level preference data infeasible, since each incorrect step in the interaction would require a corrected version. Hence, we evaluate PRMs only at the response-level and leave the step-level training and evaluation as a future work.

\begin{table}
    \centering
    \small
    \caption{Answer label performance on \textbf{GSM8K}. An FPR@0.9 Recall of 1.0 (mr: \textit{x}) means that the recall does not exceed \textit{x} without flagging everything as low confidence.}
    \begin{tabular}{ll|ccc}
        \toprule
        \textbf{Technique} & \textbf{granularity} & \textbf{AUC-ROC} ($\uparrow$) & \textbf{ECE} ($\downarrow$) & \textbf{FPR@0.9 Recall} ($\downarrow$) \\
        \midrule

         & response & 0.560 & 0.317 & 1.0 (mr: 0.15) \\
        Self-eval & step & 0.559 (-0.2\%) & 0.3125 & 1.0 (mr: 0.12) \\
        \midrule

         & response & 0.590 & 0.291 & 1.0 (mr: 0.52) \\
        Llama-3.2-11B & step & 0.669 (+13\%) & 0.159 & 1.0 (mr: 0.76) \\
        \midrule
        
         & response & \textbf{0.895} & 0.088 & 1.0 (mr: 0.88) \\
        GPT-4.1-mini & step & 0.662 (-26\%) & 0.280 & 1.0 (mr: 0.49) \\
        \midrule

         & response & 0.869 & \textbf{0.075} & 0.4385 \\
        Regression & step & 0.872 (+1\%) & 0.144 & \textbf{0.369} \\
        \midrule
        
         & response & 0.460 & 0.629 & 0.915 \\
        PRM & step & - & - & - \\
        \midrule
        
         & response & 0.658 & 0.219 & 0.773 \\
        Self-certainty & step & 0.342 (-48\%) & 0.320 & 0.958 \\
        \midrule
        
         & response & 0.605 & 0.339 & 1 (mr: 0.77) \\
        Activations & query & 0.738 (+21\%) & 0.279 & 0.655 \\
        \bottomrule
    \end{tabular}
    \label{tab:gsm-ans}
\end{table}

\subsection{White-box methods}

\textbf{Logits:}
Following \textbf{Self-certainty} \cite{self-certainty}, we compute the KL divergence of the agent's output logits from the uniform distribution as a measure of certainty. Since this approach consistently outperforms other logit-based methods, we use it as the representative logit-based white-box baseline. Self-certainty scores are normalized to fall within $[0, 1]$.

\textbf{Activations:}
Prior work \cite{activations1, activations2, activations3} suggests that hidden states of the model's final LLM layer contain information on model's behavior and can be used to extract its confidence in its response. Following this, we train a 5-layer MLP classifier on the model's final hidden states to predict a correctness score.

\section{Evaluating on private dataset}\label{app:private}

\begin{figure}
    \centering
    \includegraphics[width=0.4\linewidth]{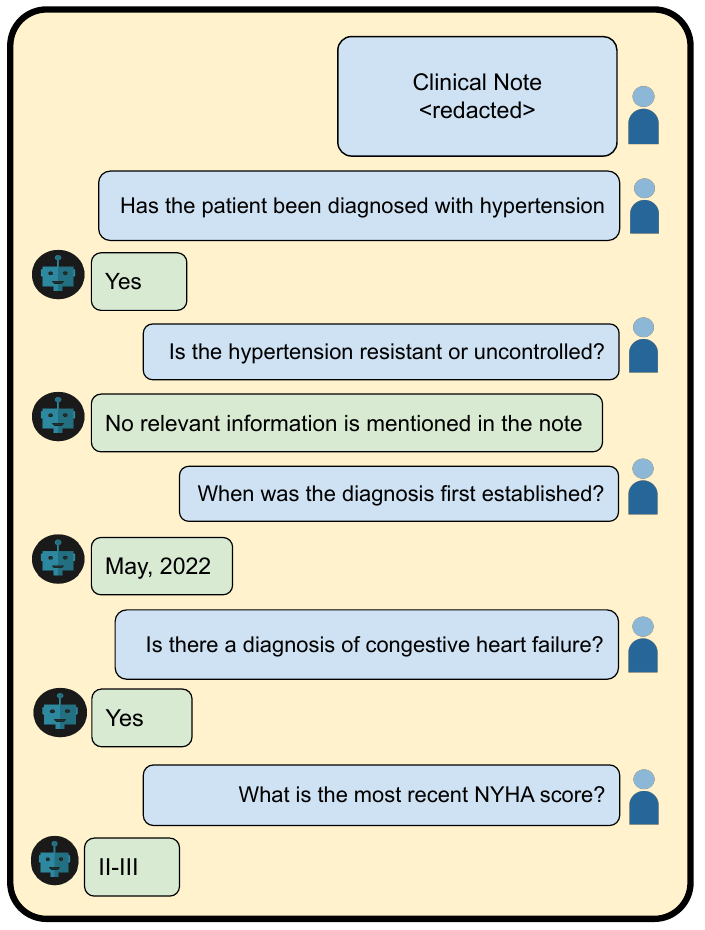}
    \caption{An example from the private clinical data.}
    \label{fig:private-example}
\end{figure}

To evaluate the applicability of confidence estimation methods in real-world settings, we tested some of the approaches on a private dataset consisting of conversational question-answering interactions over real patient clinical notes. A redacted example from this dataset is provided in Figure \ref{fig:private-example}. We do not publicly release the data due to conflict of interest as well as HIPAA compliance, and the results are therefore not reproducible. Nevertheless, it provides a valuable demonstration in a domain where trustworthiness is critical. 
Consistent with the analysis on public datasets, a regression model generating step-level scores achieves the best performance with \textbf{AUC-ROC $\mathbf{=0.940}$} and \textbf{FPR@0.9 rec $\mathbf{=0.152}$}. These results indicate that step-level confidence scoring with a regression model remains effective in complex, real-world interactions.
\end{document}